\documentclass[lettersize,journal]{IEEEtran}
\usepackage{amsmath,amsfonts}
\usepackage{algorithmic}
\usepackage{algorithm}
\usepackage{array}
\usepackage[caption=false,font=normalsize,labelfont=sf,textfont=sf]{subfig}
\usepackage{textcomp}
\usepackage{stfloats}
\usepackage{hyperref}
\usepackage{verbatim}
\usepackage{graphicx}
\usepackage{cite}
\usepackage{xcolor}
\usepackage{multirow}
\usepackage{amssymb}
\usepackage{pifont}

\begin{document}

%\title{Learning Collision-Free Trajectories for \\Robotic Pick-and-Place Tasks}
%\title{Generalizing Dynamical Movement Primitives with Reinforcement and Deep Learning}
%\title{Generalizing Dynamical Movement Primitives \\from One Demonstration}
% \title{Generalizing DMPs from One Demonstration \\ for Obstacle Avoidance Applications}
%\title{DMPI² - Generalizing Obstacle Avoidance from One Demonstration}
% \title{Learning Collision-avoidance Policies \\ using PI2}
% \title{Learning Collision-avoidance Policies using Dynamic Movement Primitives and Reinforcement Learning}
% \title{Policy Learning for Collision Avoidance Using Dynamic Movement Primitives and Reinforcement Learning}
\title{Obstacle Avoidance using Dynamic Movement Primitives and Reinforcement Learning}
%\title{Learning Obstacle-Avoidance Policies \\ with Dynamic Movement Primitives}
%\title{Automated Obstacle Avoidance with Dynamic Movement Primitives and Policy-Based Reinforcement Learning}

\author{Dominik Urbaniak$^1$, Alejandro Agostini$^2$, Pol Ramon$^1$, Jan Rosell$^1$, Ra\'ul Su\'arez$^1$, Michael Suppa$^3$
        % <-this % stops a space
\thanks{$^1$Institute of Industrial and Control Engineering, Universitat Politècnica de Catalunya, Barcelona, Catalonia, Spain (e-mail: [name].[surname]@upc.edu)}% <-this % stops a space
\thanks{$^2$Department of Computer Science, University of Innsbruck, Innsbruck, Austria (e-mail: alejandro.agostini@uibk.ac.at)}% <-this % stops a space
\thanks{$^3$Roboception GmbH, Munich, Germany (e-mail: michael.suppa@roboception.de)}
\thanks{This work was funded by the European Union’s Horizon 2020 research and innovation programme under the Marie Skłodowska - Curie grant agreement No 956670 and by the Austrian Science Fund (FWF) Project P36965 [DOI: 10.55776/P36965].}

\thanks{Manuscript received MONTH XX, XXXX; revised MONTH XX, XXXX.}}

% The paper headers
%\markboth{IEEE ROBOTICS AND AUTOMATION LETTERS, VOL. X, NO. X, MONTH 2025 }%
%{Shell \MakeLowercase{\textit{et al.}}: A Sample Article Using IEEEtran.cls for IEEE Journals}

%\IEEEpubid{0000--0000/00\$00.00~\copyright~2021 IEEE}
% Remember, if you use this you must call \IEEEpubidadjcol in the second
% column for its text to clear the IEEEpubid mark.

\maketitle

\begin{abstract}
Learning-based motion planning can quickly generate near-optimal trajectories. However, it often requires either large training datasets or costly collection of human demonstrations. This work proposes an alternative approach that quickly generates smooth, near-optimal collision-free 3D Cartesian trajectories from a single artificial demonstration. The demonstration is encoded as a Dynamic Movement Primitive (DMP) and iteratively reshaped using policy-based reinforcement learning to create a diverse trajectory dataset for varying obstacle configurations. This dataset is used to train a neural network that takes as inputs the task parameters describing the obstacle dimensions and location, derived automatically from a point cloud, and outputs the DMP parameters that generate the trajectory. The approach is validated in simulation and real-robot experiments, outperforming a RRT-Connect baseline in terms of computation and execution time, as well as trajectory length, while supporting multi-modal trajectory generation for different obstacle geometries and end-effector dimensions. Videos and the implementation code are available at \url{https://github.com/DominikUrbaniak/obst-avoid-dmp-pi2}.

%for varying obstacles and end-effector dimensions using Dynamic Movement Primitives (DMP). The parameters of one DMP are initialized with a single artificially-generated, linear, minimum-jerk demonstration and, subsequently, reshaped using policy-based reinforcement learning to generate a dataset for different obstacle scenarios. This dataset is used to train a neural network that takes as inputs the task parameters describing the obstacle dimensions and location, derived automatically from a point cloud, and outputs the DMP parameters that generate the trajectory. We demonstrate the validity of the approach in simulated and real scenarios for variable complexity tasks. Videos and the implementation code are provided at \url{https://github.com/DominikUrbaniak/obst-avoid-dmp-pi2}.
\end{abstract}
% between the initial and goal positions of the movement
%However, in many scenarios, such collisions can be avoided by curved trajectories that consider the end-effector dimensions. 
%Policy Improvement with Path Integrals (PI²)

%Linear motion can result in collisions between the robot end-effector and other objects. Autonomous obstacle avoidance approaches are often either computationally expensive, suboptimal or jerky. 

\section{Introduction}
%\PARstart{C}{urrently}
A motion planner for autonomous robotic manipulation should be able to quickly generate smooth optimal trajectories in different scenarios~\cite{tamizi2023review}. Sampling-based motion planners often struggle to quickly find near-optimal trajectories due to frequent online resampling~\cite{kuffner2000rrt,karaman2011sampling}. Learning-based methods shift the major computational effort to offline training, significantly reducing the time to find near-optimal solutions online~\cite{qureshi2020motion,fishman2023motion}. More data-efficient learning-based approaches use movement primitives to quickly reproduce and generalize human demonstrations~\cite{pervez2018learning,zhou2020movement,tekden2023neural,zhang2024using,wang2025pi2,huang2025task}. However, generalization capabilities, smoothness, and optimality depend on the number and quality of the human demonstrations, which are expensive to generate. 

\begin{figure}[t]
    \centering
    \includegraphics[width=0.49\textwidth]{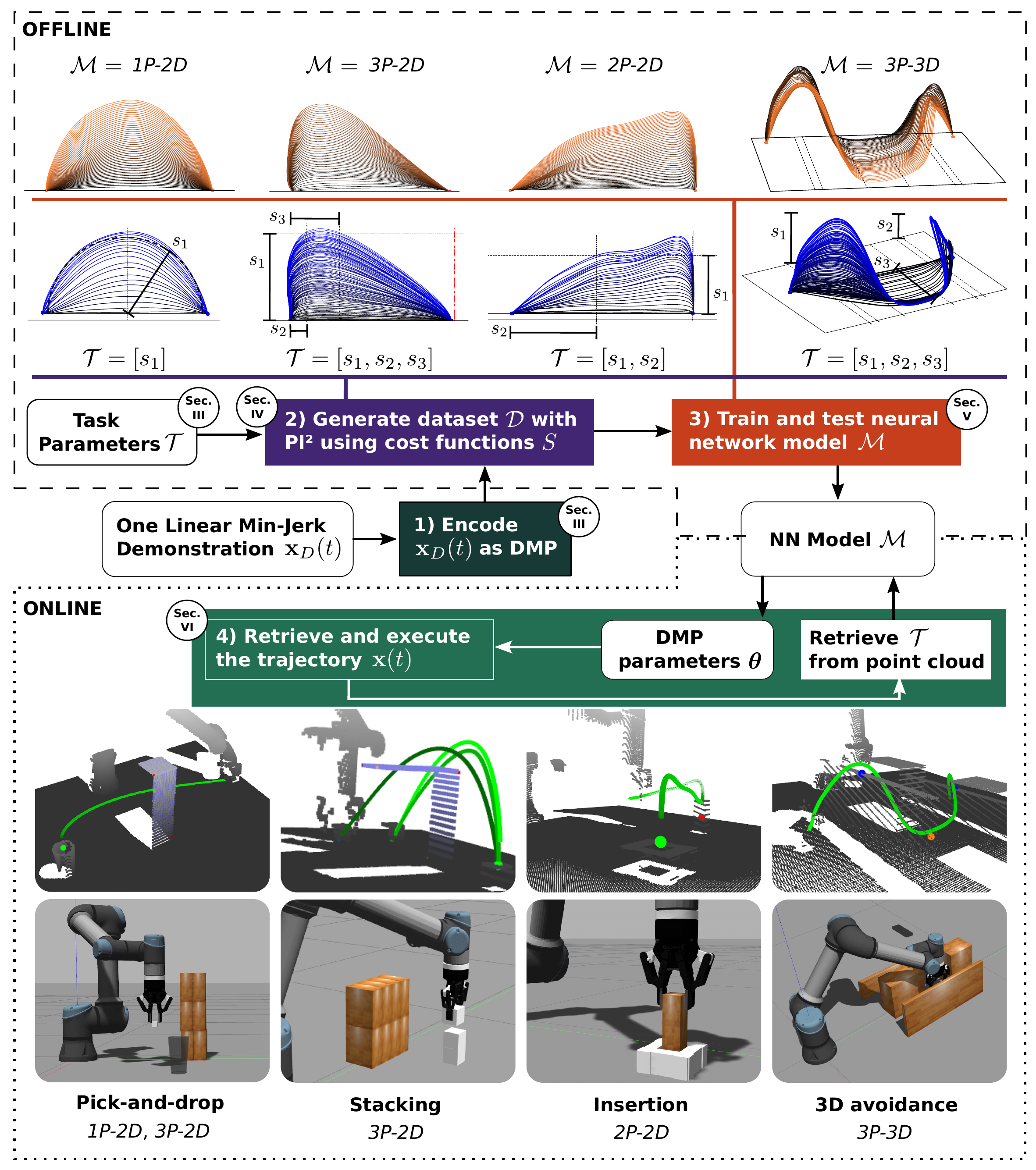}
    \caption{System overview including steps 1) to 4) shown as colored blocks, and four example models and applications.}
    \label{fig:fig1}
\end{figure}

In this work, we address these limitations by using only a single artificial demonstration, which is automatically adjusted to avoid evolving obstacle configurations, generating a dataset of optimal obstacle avoidance trajectories for different scenarios. After mapping the dataset to a small neural network, a near-optimal trajectory can be quickly generated in an unseen scenario. Trajectory smoothness is facilitated by the minimum-jerk demonstration and acceleration and jerk penalties during the dataset generation. Figure~\ref{fig:fig1} provides an overview of the proposed system that efficiently learns trajectory distributions offline, and automatically retrieves a suitable trajectory online: 
%proposing a methodology that efficiently learns trajectory distributions offline, and automatically retrieves solutions for different scenarios online. Figure~\ref{fig:fig1} provides an overview of the approach: 
first, the demonstration $\textbf{x}_D(t)$ is encoded as a Dynamic Movement Primitive (DMP)~\cite{ijspeert2013dynamical, ginesi2021overcoming} (\mbox{block 1}), offering inherent generalization to different start and goal positions, as well as rotations and scales.
%first, a single artificially-generated minimum-jerk demonstration $\textbf{x}_D(t)$ is encoded as a Dynamic Movement Primitive (DMP)~\cite{ijspeert2013dynamical, ginesi2021overcoming} (block 1), offering inherent generalization to different start and goal positions, as well as rotations and scales. 
%$\textbf{x}_D(t)$  $\mathcal{T}$ $n=3$ $\mathcal{D}$ $\mathcal{M}$ $\boldsymbol{\theta}$  $\textbf{x}(t)$
%: From one DMP-encoded demonstration (1), various trajectory shapes can be generated with PI² (2) and mapped by a neural network offline (3), allowing quick trajectory retrieval online (4). Task parameters describe the obstacle configuration and are derived from point clouds.
During the offline training, a trajectory dataset $\mathcal{D}$ is generated by iteratively perturbing the initial demonstration based on cost functions $S$, using Policy Improvement with Path Integrals (PI²)~\cite{theodorou2010learning}, a policy-based Reinforcement Learning (RL) algorithm (\mbox{block 2}). At each iteration, the cost functions define the optimal collision-free trajectory for a unique obstacle configuration, which are described by up to three task parameters $\mathcal{T}$ in two or three dimensional Cartesian space. The dataset is mapped to a neural network model $\mathcal{M}$, where the number of task parameters $n$ and Cartesian dimensions $d$ are represented as $n$\textit{P}\textit{-}$d$\textit{D} (block 3). Online, the task parameters are automatically derived from a point cloud and used to infer DMP parameters $\boldsymbol{\theta}$ that allow to quickly generate a near-optimal collision-free trajectory $\textbf{x}(t)$ in a new scenario (block 4). This methodology allows learning different models for varying avoidance applications. 

The main contributions of this work comprise:
\begin{itemize}
    \item A methodology for fast, smooth and near-optimal trajectory generation from different task parameters, after offline training using PI² for generating the dataset, including an automatic derivation of the task parameters from a point cloud.
    \item Obstacle avoidance with three continuous task parameters that can consider different obstacle configurations and end-effector dimensions, and find multi-modal solutions.
    \item A comprehensive experimental evaluation, including comparisons with the RRT-Connect planner and a linear baseline in terms of computation time, execution time, and trajectory length.
    % Comprehensive experimental evaluation including a comparison with the RRT-Connect planner and a linear baseline regarding computation and execution times, and trajectory lengths. 
\end{itemize}
%It extends our previous work~\cite{urbaniak2021combining}. 

After this introduction, Section~\ref{sec:related_work} presents the related work. Then, the preliminaries are introduced in Section~\ref{sec:preliminaries}. Sections~\ref{sec:pi2} and~\ref{sec:nn} describe the offline training methodology. Section~\ref{sec:sim} describes the simulated and real experiments, including the task parameter derivation from point clouds. Finally, the performance and future directions are discussed in Section ~\ref{sec:discussion}.

%: first, the data generation with PI² (Sec.~\ref{sec:pi2}), then the learning of the neural network models (Sec.~\ref{sec:nn}). .   

%defined task parameters, a neural network model   One linear minimum-jerk demonstration is parameterized as a Dynamical Movement Primitive (DMP). These parameters are perturbed based on different cost functions using policy-based Reinforcement Learning (RL). Each iteration, the trajectory parameters are collected together with the cost serving as training data and labels for training a neural network using Deep Learning (DL). Depending on the complexity of the trajectory and the number of task parameters that define the trajectory shape, training takes a few minutes to a few hours. Finally, a near optimal collision-free trajectory can be generated in less than $0.1$ s after retrieving the required task parameters from a point cloud.

\section{Related Work}
\label{sec:related_work}
Obstacle avoidance is a fundamental challenge in robotic motion planning and is approached with various methods~\cite{lavalle2006planning}. The RRT-Connect planner~\cite{kuffner2000rrt} samples random configurations from the start and goal positions online. It is fast and probabilistically complete, however, the solutions are not optimal. Asymptotically optimal planners (e.g. RRT*)~\cite{karaman2011sampling} come at the cost of longer computation times~\cite{orthey2023sampling}. A recent approach combines the strengths of both methods using heuristics and pruning strategies~\cite{he2025irrtc}. 

Learning-based approaches present a promising alternative~\cite{tamizi2023review}, leveraging offline training to reduce the computational effort online. Motion Planning Networks~\cite{qureshi2020motion} use training data generated by RRT* to learn to predict the next collision-free robot configuration given a point cloud and the current and goal configuration. 
%In a pick-and-place experiment, unseen scenarios with different start and goal positions can be handled. 
%Our approach benefits from the DMP's roto-dilatation invariance that handles different start and goal positions inherently. 
%Motion Policy Networks~\cite{fishman2023motion} utilize millions of simulated training data samples to achieve smooth collision-free motion from a point cloud in various environments, also showing sim-to-real transfer. 
Other learning-based approaches encode one or multiple demonstrations as movement primitives (MP)~\cite{pervez2018learning,zhou2020movement,tekden2023neural,zhang2024using,wang2025pi2,huang2025task}, which provide some generalization ability by design. Further generalization capabilities are often achieved by specified task parameters, which describe how the demonstrations vary, e.g. by different goal positions, via-points or obstacle dimensions. 
%ude2010task,zhou2017task,ridge2020training,akbulut2021acnmp,,yildirim2024conditional
%,perez2023learning
%two demonstrations that avoid an obstacle with two different heights. The final model generates the forcing term for a DMP, showing consistent interpolation and extrapolation abilities. For a sweeping experiment, four demonstrations with different via-points are collected. In the application, the via-points are defined by the $X-Y$ positions of the trash object which are detected automatically by a camera using fiducial markers.  
A mixture of Gaussian Mixture Models (GMMs) is trained on a few demonstrations to generate the forcing term of a DMP, showing consistent interpolation and extrapolation abilities in an obstacle avoidance and sweeping task with up to two task parameters~\cite{pervez2018learning}. The position of the object in the sweeping task is detected automatically by a camera using fiducial markers. By contrast, our work generalizes with up to three task parameters plus random variations of the start and goal positions, with object detection and task parameter derivation directly from a point cloud.

Task parameters can also be detected indirectly from scene images. Demonstrations consisting of images paired with robot trajectories are mapped into scene-motion embeddings using neural fields~\cite{tekden2023neural}. In pick-and-drop, insertion and obstacle avoidance experiments, trajectories are generated for unseen configurations with up to three interpolated task parameters after optimizing the scene reconstruction loss. Our approach can be applied to similar experiments without requiring additional online optimization.
%The number of required demonstrations is $3^n$ depending on the number of task parameters $n$. 

Another approach trains a neural network to map task parameters to the GMM of a via-point MP~\cite{zhou2020movement}, which is related to DMPs. In a 2D obstacle avoidance experiment with three circles, human demonstrations are collected to learn collision-free paths from randomly placed start, goal and obstacle positions. The trained model shows the ability to consider multi-modal solutions, going around the obstacles to the left or right, or straight through them. In another experiment, fifth-order polynomials are generated with defined, uniformly-sampled end-velocities. This automated generation allows to ensure a consistent quality of the demonstration. Our work relies on such an artificially-generated demonstration and can find multi-modal solutions, while also enabling automatic task parameter derivation from a point cloud. 

Alternatively, PI² can be used to improve the quality of parameterized human demonstrations by optimizing user-defined cost functions. 
For example, stability and acceleration penalties are used to improve the grasp quality~\cite{stulp2012reinforcement}. It is also used to adapt a demonstration to new scenarios. For a polishing task, a highly accurate trajectory is learned, which passes through multiple via-points, reaches a specified goal location and keeps a desired contact force~\cite{wang2025pi2}. In these works, the PI² optimization must be run online before every new scenario. On the contrary, our work utilizes the PI² optimization offline to generate a dataset, allowing a neural network to infer DMP forcing term parameters in real-time. 

Instead of optimizing the DMP parameters, other approaches use PI² to optimize the parameters of potential fields for obstacle avoidance tasks~\cite{huang2025task}. This allows keeping the same parameters for similar problems. The approach is designed for obstacle avoidance with a mobile robot that considers the object dimensions in the potential field generation and an additional safety distance. In their cost function, the authors compute the similarity to the demonstration, which they compare to solutions by sampling-based motion planners. Our contribution compares well to RRT-Connect~\cite{kuffner2000rrt} regarding execution times and trajectory lengths, but without the computational burden inherent to deliberative approaches.
% It can consider different end-effector dimensions and safety distances without re-training.
%
%Additionally, they include accelerations and velocities to encourage the generation of a smooth trajectory. 

%In our previous work~\cite{urbaniak2021combining}, PI² was used to perturb the shape of a linear min-jerk demonstration. The cost function is designed to continuously increase the curvature of the linear demonstration while keeping the trajectory in a defined scope. At each iteration of the PI² optimization, the intermediate DMP parameters are collected in a dataset together with the corresponding task parameters which are defined by the continuous cost and initial conditions. A neural network is trained to allow quick online regression of the DMP parameters based on two given task parameters that describe an obstacle of different sizes located in the center between start and goal. The generalization to different goal positions is achieved automatically by transforming the DMP parameters utilizing their roto-dilatation invariance~\cite{ginesi2021overcoming}. 
%

In this work, we extend our previous contribution~\cite{urbaniak2021combining} by using up to three continuous task parameters that enrich obstacle avoidance capabilities. These task parameters are automatically detected from a point cloud, rather than handcrafted. Additionally, we are able to generate multi-modal solutions using offsets that ensure obstacle avoidance for different end-effector dimensions. We use the more robust DMP formulation by~\cite{ginesi2021overcoming}, which allows replacing the goal accuracy cost function by a cost that penalizes the initial acceleration and jerk to improve the motion smoothness. Finally, we carry out a more comprehensive experimental evaluation, including four different and more complex scenarios, a comparison with RRT-Connect, and real-robot experiments. 

\section{Preliminaries}
\label{sec:preliminaries}
%DMPs are used to represent the robot motion, initialized by a single demonstration. Subsequently, PI² iteratively perturbs the initial DMP parameters, creating various trajectory shapes that avoid different obstacles geometries, described by the task parameters. 
%The neural network learns to map $\mathcal{T}$ to DMP parameters $\boldsymbol{\theta}$, which allows to quickly generate a collision-free trajectory during the online application.
\subsection{Task Parameterization}
%The task parameters $\mathcal{T}$ describe different obstacle avoidance scenarios. 
Figure~\ref{fig:fig1} shows four example models $\mathcal{M}$ (\textit{1P-2D}, \mbox{\textit{3P-2D}}, \textit{2P-2D}, \textit{3P-3D}). The task parameters $\mathcal{T}=[s_1,\dots,s_n]$ describe different obstacle sizes and locations.    
The simple \textit{1P-2D} uses one task parameter for a symmetric obstacle between the start and goal. In contrast, the more complex \textit{3P-2D} model can capture the size and location of the obstacle more precisely, serving as the main example in this work.
The other two models are designed for more specific problems. The \textit{2P-2D} model considers an obstacle that surrounds the goal (insertion). The \textit{3P-3D} model creates 3D trajectories for a specific avoidance problem with two walls and a gap. New models for other applications can be created by following the methodology explained in this work.%Table~\ref{tab:experimental_variations2} compares the offline training metrics of the four models.
%While these three models create 2D trajectories (along $\mathbf{e}_1$ and $\mathbf{e}_2$), t
%(avoidance in $\mathbf{e}_2$)  (avoidance in $\mathbf{e}_3$)
\subsection{Dynamic Movement Primitives}
DMPs~\cite{ijspeert2013dynamical} model each degree of freedom as a spring-damper system with a learnable forcing term $f(\phi)$,

\begin{equation}
\begin{aligned}
\tau \dot v &= K\,(g - x) - D\,v + (g - x_0)\,f(\phi)\\
\tau \dot x &= v,
\end{aligned}
\end{equation}
where $\tau$ is a temporal scaling factor, $x_0$ and $x$ are the initial and current positions, $v$ the current velocity and $g$ the goal position. The spring and damping constants $K=25$ and \mbox{$D=2\sqrt{K}$} make the system critically damped. The phase variable $\phi\in(0,1]$ decays exponentially, 
\begin{equation}
\tau \dot \phi = -\,\alpha\,\phi,
\end{equation}
where $\alpha$ is a constant that determines the exponential decay.
Defining $f(\phi)$ as a normalized linear combination of basis functions allows reshaping the linear behavior arbitrarily and imitating a demonstrated trajectory.
Then, $f(\phi)$ is approximated with $N$ Gaussian radial basis functions $\psi_i(\phi)$,
\begin{equation}
\label{eq:dmp_1d}
f(\phi)=\frac{\sum_{i=0}^{N}\theta_i\,\psi_i(\phi)}
           {\sum_{i=0}^{N}\psi_i(\phi)}\,\phi, \; \text{with} \;%i=0,\dots,N
           \psi_i(\phi)=\exp\bigl(-\,h_i\,(\phi-c_i)^2\bigr),
\end{equation}
where $\theta_i$ are the \textit{DMP parameters} or \textit{forcing term parameters}, which are placed at centers $c_i$ with widths $h_i$ along the trajectory with $T$ time steps,
\begin{equation}
c_i=\exp\Bigl(-\,\alpha\,\frac{i\,T}{N}\Bigr),\;
h_i = \frac{\tilde h}{(c_{i+1}-c_i)^2}, %\quad \\ 
\; h_N = h_{N-1},
\end{equation}

\noindent where $\tilde{h}=0.5$ defines the overlap of the basis functions. The initial forcing term parameters can be derived from a demonstrated trajectory by solving Equation~(\ref{eq:dmp_1d}).
For \mbox{$d$-dimensional} motion, one DMP per dimension is used sharing the phase $\phi$. The formulation
%To improve the spatial scaling of multi-dimensional DMPs, Equation~(\ref{eq:dmp_1d}) is reformulated such that all DMPs share the same phase but the forcing term is no longer scaled by $\textbf{g}-\textbf{x}_0$,
\begin{equation}
\begin{aligned}
\tau \dot{\textbf{v}}&=\textbf{K}\,(\textbf{g}-\textbf{x})
                     -\textbf{D}\,\textbf{v}
                     -\textbf{K}\,(\textbf{g}-\textbf{x}_{0})\,\phi
                     +\textbf{K}\,\textbf{f}(\phi),\\
\tau \dot{\textbf{x}}&=\textbf{v},
\end{aligned}
\end{equation}
preserves the trajectory shape under arbitrary scalings and rotations, where $\textbf{x},\textbf{x}_0,\textbf{v},\textbf{g},\textbf{f}(\phi) \in  \mathbb{R}^d$ and $\textbf{K},\textbf{D} \in \mathbb{R}^{d\times d}$ are diagonal matrices.

To initialize the trajectory, a one-dimensional demonstration $\textbf{x}_D(t)$ is artificially generated by a fifth-order polynomial that starts and ends with zero velocity, resulting in a linear minimum jerk trajectory, modeled to imitate human voluntary arm movements~\cite{flash1985coordination}. This demonstration is applied to the primary axis $\mathbf{e}_1$ of a three-dimensional DMP,
\begin{equation}
\mathbf{e}_1 = \frac{\textbf{g} - \textbf{x}_0}{L},
\end{equation}
where $L=\lVert \textbf{g} - \textbf{x}_0 \rVert$ and the remaining basis vectors $\mathbf{e}_2$, $\mathbf{e}_3$ are orthogonal $\mathbf{e}_1$ to span the 3D Cartesian space. Axis $\mathbf{e}_1$ is always along the direction from the start to the goal positions of the trajectory.
%These multi-dimensional DMPs can preserve the initialized trajectory shape to arbitrary rotation and scaling by applying the same transformation to the forcing term that maps the original vector $\textbf{g}-\textbf{x}_0$ to the new start and goal configuration. 

Moreover, this representation allows generating multi-modal 2D trajectories by rotating the learned forcing term around the primary axis $\mathbf{e}_1$ using an angle $\beta$, 
\begin{equation}
    \boldsymbol{\theta}_{new} =
\begin{bmatrix}
\boldsymbol{\theta}_{\mathbf{e}_1} \\
\cos(\beta) \boldsymbol{\theta}_{\mathbf{e}_2} \\
\sin(\beta) \boldsymbol{\theta}_{\mathbf{e}_2}.
\end{bmatrix}
\end{equation}

More details on the DMP formulation are provided in~\cite{ginesi2021overcoming} along its implementation as a Python library that is used in this work.

\subsection{Policy Improvement with Path Integrals}
PI²~\cite{theodorou2010learning} is a model-free, policy-based RL algorithm that optimizes the DMP parameters $\theta^j_i$ at each iteration $j$, creating \mbox{$q=1,...,Q$} perturbed policies,  
%each by creating $M$ trajectory samples using random exploration noise $\epsilon$ based on stochastic sampling . A simplified version~\cite{stulp2012policy} samples $\epsilon_i$ only once at the first time step, and 2) postpone the policy update until after the last time step

%is derived from stochastic optimal control (SOC) and that remains numerically robust even in high-dimensional control problems~\cite{theodorou2010learning}. Given the parameters of a policy, e.g. the learned parameters $\theta_{i,D}$ of a DMP, PI² iteratively optimizes the parameters with respect to user-specified cost functions. At each iteration, the current policy is perturbed by a random exploration noise $\epsilon_{i,t}$, originally sampled at each time step from a Gaussian distribution $\mathcal{N}(0,\sigma^2)$, where $\sigma^2$ is the only PI² hyper-parameter that must be tuned. Stulp et al.~\cite{stulp2012policy} proposed two simplifications: 1) sample $\epsilon_i$ only once at the first time step, and 2) postpone the policy update until after the last time step, using the accumulated cost. These changes improve the performance, reduce the convergence time and don't violate the any assumptions from the SOC derivation~\cite{stulp2012policy}.  
\begin{equation}
\theta_{i,q}^{j+1}= \theta_{i}^{j}+\epsilon_{i,q}^{j},\qquad
\epsilon_{i,q}^{j}\sim\mathcal N(0,\sigma_i^2).
\end{equation}
Each policy is evaluated by a trajectory‑level cost $S(\theta_{i,q}^{j})$ (discussed in Section~\ref{sec:pi2}) and assigned an importance weight $W(\theta_{i,q}^j)$,
\begin{equation}
\label{equation_pi2_alt}
W(\theta_{i,q}^{j})=
\exp\Bigl(
-\gamma,
\frac{S(\theta_{i,q}^{j})-\min S(\boldsymbol{\theta}_{i}^{j})}
{\max S(\boldsymbol{\theta}_{i}^{j})-\min S(\boldsymbol{\theta}_{i}^{j})}
\Bigr),
\end{equation}
The updated DMP parameters are the normalized, weighted average of all $Q$ samples,
\begin{equation}
\theta_{i}^{j+1}=
\frac{\sum_{q=1}^{Q} W(\theta_{i,q}^{j}),\theta_{i,q}^{j}}
{\sum_{q=1}^{Q} W(\theta_{i,q}^{j})}.
\end{equation}
This simplified PI² formulation improves the performance and accelerates convergence by sampling $\epsilon^j_{i,q}$ only once, and delaying the policy update until after the last time step, using the accumulated cost~\cite{stulp2012policy}.
%and updates  with the accumulated cost $S(\theta_{i,m}^{j})$
We refer below to the set of DMP parameters $\theta_i$ as the vector $\boldsymbol{\theta}$. 

Sampling all $\epsilon_i$ from the same distribution results in stronger trajectory shape perturbations in the beginning of the trajectory, due to the exponentially decaying phase $\phi$. To encourage a symmetric shape perturbation, in this work, the exploration noise $\epsilon_i$ is sampled from a Gaussian distribution $\mathcal{N}(0,\sigma_i^2)$, where
\begin{align}
\sigma_i^{2} &=
  \bigl(\exp(\hat{\sigma}_i) - 1\bigr)^{2} \\
\hat{\sigma}_i &=
  \sigma^- +
  (\sigma^+-\sigma^-)
  \left(\frac{i}{N-1}\right)^2,
  %&& i = 0,\ldots,N-1,
\end{align}
and $\sigma^-,\sigma^+$ are two empirically chosen hyper-parameters that create an exponential growth of $\sigma_i^2$.

\section{Data Generation with PI²}
\label{sec:pi2}
The PI² optimization generates a training dataset $\mathcal{D}$ for a task that can be parameterized by $\mathcal{T} \in \mathbb{R}^n$. 
A single PI² optimization run generates data along one continuous dimension $s_1$. This is shown in Fig.~\ref{fig:fig1} for all four examples where the optimization starts with the linear demonstration, which develops continuously along one dimension (black-to-blue gradient). More discrete task dimensions can be added by performing one PI² optimization run for each discretized combination of parameters. This scales exponentially with each discretized dimension~\cite{tekden2023neural}. As an alternative, we show for the \textit{3P-2D} model that the dataset can be generated from randomly initialized values. 

%Whether predefined or randomly initialized, a unique task parameter set $\mathcal{T}_m = \{S_{shape},p^{(1)}_m,...\} \in \mathbb{R}^n$ is optimized during one PI² optimization run where $M$ is the number of PI² optimization runs for one parameterized task. Each optimization run starts at $j=0$ with the forcing terms of the DMP set to $\boldsymbol{\theta}_0$, which is reproducing the linear minimum-jerk demonstration. One iteration of the optimization comprises i) sampling $K$ variations $\boldsymbol{\theta}'_k$ using a multivariate normal distribution, ii) computing the DMP trajectories $\textbf{x}'_k$ and evaluating their costs $S'_k$, iii) creating the new forcing term parameters $\boldsymbol{\theta}_{j+1}$ as weighted sum based on the costs $S'_k$, and iv) computing and assessing the new trajectory $\textbf{x}_{j+1}$. At each iteration $j$, the new parameters $\boldsymbol{\theta}_{j+1}$ are added to the dataset $\mathcal{D}_m$ as label or target. The cost $S_{shape,j+1}$ is added as corresponding input. In case of $n>1$ (which requires $M>1$), the additional task parameters $p^{n-1}_m$ are also added as input parameters to $\mathcal{D}_m$ which are constant during one optimization run $m$.     

\begin{figure}[t]
    \centering 
    \includegraphics[width=0.49\textwidth]{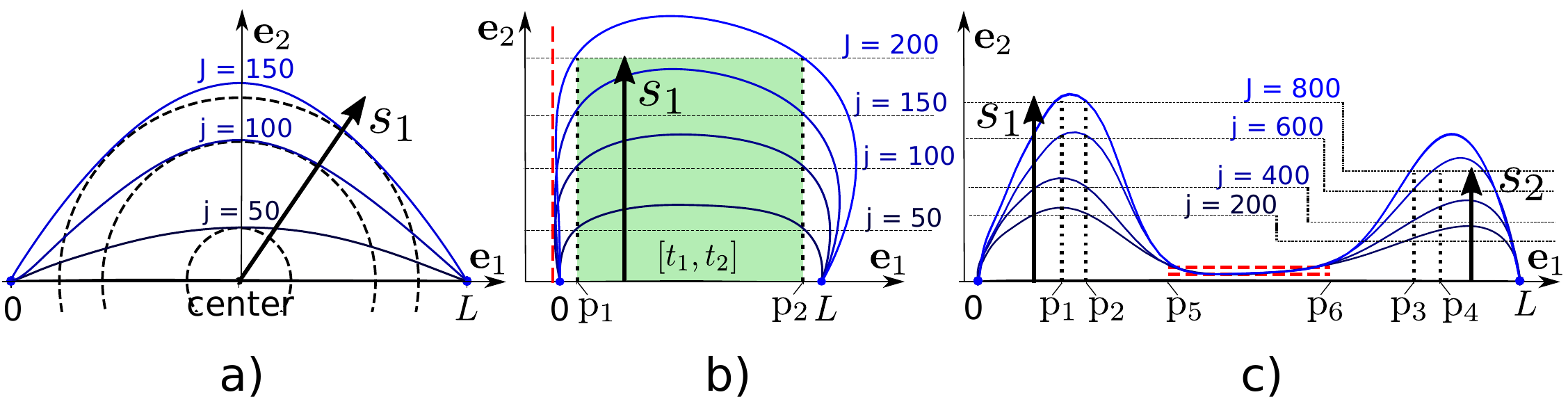}
    \caption{PI² cost function effects: Continuous $S_{shape}$ cost function, a) for one task parameter, and b-c) for three task parameters. $S_{scope}$ constrains the trajectory to the left in b) and within a gap in c). Also in c), task parameter $s_2$ is defined by a constant ratio $s_1/s_2$.}
    \label{fig:cost_function_shape}
\end{figure}

%\subsection{Cost functions}
The PI² optimization is driven by a set of cost functions $S$. In this work, we use three different types of cost functions: i) $S_{shape}$ to continuously reshape the trajectory, ii) $S_{scope}$ to constrain the reshaping process to desired areas, and, iii) $S_{acc,0}$ to penalize the initial acceleration and $S_{jerk}$ to penalize the overall jerk. All cost functions are accumulated. Different scaling factors $C$ allow prioritizing the influence of a cost function term.
%Compared to our previous work~\cite{urbaniak2021combining}, the acceleration and jerk costs are new and the cost function on the goal accuracy is omitted, as the utilized DMP library~\cite{ginesi2021overcoming} ensures the goal convergence by definition, given a desired accuracy target.
%Give examples of the impact of the cost functions, with / without S_{shape}, S_{scope}, S_{acc,0}, S_{jerk}
%
%\subsection{Continuous perturbation of the trajectory shape}
The cost $S_{shape}$ continuously decreases, creating optimal trajectories for different scenarios, 
\begin{equation}\label{eq:cost_shape}
    S_{shape} = -C_{shape}\min_{t \in [t_1, t_2]} \mathbf{u}(t),
\end{equation}
where $C_{shape}=1$ throughout this work, $t_1$ and $t_2$ define a section of the trajectory and $\mathbf{u}(t)$ is a one dimensional vector of positions or distances. Figure \ref{fig:cost_function_shape} illustrates three examples that keep decreasing over time until a target value $S^*_{shape}$ is reached, which stops the optimization. In a), $S_{shape}$ is defined as a circle with maximum radius that fits inside the trajectory placed at the center point. This is used for the \textit{1P-2D} model. In b), $S_{shape}$ is defined as the minimum $\mathbf{e}_2$ value of the trajectory between $p_1$ and $p_2$. This function is used for the \textit{3P-2D} and \textit{2P-2D} models, as well as for the third dimension ($\mathbf{e}_3$) of the \textit{3P-3D} model. A more complex scenario is shown in c), where the $\mathbf{e}_2$ value of the trajectory is evaluated at four points ($p_{1},...,p_4$), representing two trajectory peaks with a constant height ratio $s_1/s_2$. This function is used for the second dimension of the \textit{3P-3D} model.   
%, the corresponding $\mathbf{e}_1$ locations at $t_1$ and $t_2$, respectively

%\subsection{Constraining the trajectory shape}
The second type of cost function is used to restrain the trajectory to pass barriers that are defined by $S_{scope}$ as
\begin{equation}\label{eq:cost_scope}
    S_{scope} = -C_{scope}\sum_{t=t_1}^{t_2} \min(0,\eta(\boldsymbol{\nu}(t)-\hat{v})+m),
\end{equation}
where $C_{scope}=1$ unless specified differently, $\boldsymbol{\nu}(t)$ is a one-dimensional vector of positions, $\hat{v}$ defines a reference that $\boldsymbol{\nu}(t)$ is compared to, considering an offset $m$, and $\eta \in \{-1,1\}$ defines either a upper or lower bound. The effect can be seen in Fig.~\ref{fig:cost_function_shape}b, where $\boldsymbol{\nu}(t)=\textbf{x}_{\mathbf{e}_1}(t)$, $\hat{v}=0$, $m=0.02L$ and $\eta=1$. Again, c) shows a more complex scenario with two instances of $S_{scope}$ to constrain the trajectory between $p_5$ and $p_6$ with an upper ($\eta = -1$) and lower ($\eta=1$) bound, where $\boldsymbol{\nu}(t)=\textbf{x}_{\mathbf{e}_2}(t)$ for $t\in [t_5,t_6]$, $\hat{v}=0.02L$ and $m=0.007L$. %Note, that this cost function only penalizes the

%\subsubsection{Penalizing acceleration and jerk}
The optimization with PI² also allows considering features other than the trajectory positions. Particularly within collaborative robotics, the psychological safety of humans can be improved by avoiding sudden accelerations of the robot~\cite{lasota2017survey}. To this end, the two following cost functions penalize the initial acceleration and overall jerk. The cost for the initial acceleration $S_{acc,0}$ is defined as 
%\frac{1}{100}
\begin{equation}\label{eq:cost_acc}
    S_{acc,0} =  C_{acc,0}\sum \left| \ddot{\textbf{x}}(0) \right|, 
\end{equation}
where $\ddot{\textbf{x}}(0)$ is the initial acceleration of a multi-dimensional trajectory $\textbf{x}(t)$.
The cost for the overall jerk $S_{jerk}$ is defined as
\begin{equation}\label{eq:cost_jerk}
    S_{jerk} = C_{jerk}\sqrt{\sum_{t=1}^{T-1} \left( \ddot{\textbf{x}}(t+1) - \ddot{\textbf{x}}(t) \right)^2},
\end{equation}
where $C_{acc,0}=0.01$ and $C_{jerk}=0.05$ are scaling constants empirically determined to ensure that $S_{shape}$ has priority, as $S_{shape}$ encourages longer trajectories leading to larger acceleration and jerk.
%\subsubsection{Weighted sum of cost functions}
%When combining different cost functions, the order of priority can be defined by a weighted sum. It is required when two cost functions counteract each other. $S_{shape}$ encourages longer trajectories leading to larger acceleration and jerk.  $S_{scope}$ is supposed to constrain the trajectory to the desired regions, leading to zero costs. %As shown in Fig.~\ref{fig:cost_function_shape}, more complex trajectories can be generated with a larger set of cost functions. However, the optimization time increases.

%\subsection{1P-2D model}
%Three cost functions are deployed for this model. The $S_{shape}$ in Equation~\eqref{eq:cost_shape} features $\mathbf{u}(t)$ as a vector of Euclidean distances of all points in the trajectory to the center point using the \textit{cdist} function of scipy~\cite{virtanen2020scipy}. As the start and goal of the trajectory are fixed, there is a theoretical limit with $s_1<0.5L$. The optimization target is set empirically as $S^*_{shape}=-0.47L$ to avoid slow convergence to that limit. Additionally, a $S^{bottom}_{scope}$ cost (Eq. ~\eqref{eq:cost_scope}) is added with $\boldsymbol{\nu}(t) = \textbf{x}_{\mathbf{e}_2}(t)$ and $\hat{v}=m=0$, penalizing any $\mathbf{e}_2<0$ of trajectories that would cause a collision with the ground. To emphasize the importance of this cost, it is scaled by a factor $C^{bottom}_{scope}=10$.
%\subsection{3P-2D model}
As an example, five cost functions are deployed for the \mbox{\textit{3P-2D}} model. The $S_{shape}$~\eqref{eq:cost_shape} features $\mathbf{u}(t)=x_{\mathbf{e}_2}([t_1,t_2])$, the $\mathbf{e}_2$ components of the trajectory between $p_1$ and $p_2$. At the beginning of each PI² optimization, $p_1$ and $p_2$ are sampled randomly within $[0.03L, 0.97L]$, with the condition $p_2 \geq p_1$. Additionally, three $S_{scope}$ cost functions~\eqref{eq:cost_scope} are added. The first is $S^{bottom}_{scope}$ with $\boldsymbol{\nu}_1(t) = \textbf{x}_{\mathbf{e}_2}(t)$, $\hat{v}_1=m_1=0$ and $\eta_1=1$, penalizing any $\mathbf{e}_2<0$ of trajectories that would cause a collision with the ground. The other two constrain the trajectory's $\mathbf{e}_1$ coordinates, with $\boldsymbol{\nu}_{23}(t) = \textbf{x}_{\mathbf{e}_1}(t)$, $\hat{v}_2=0$, $\eta_2=1$ and $m_{23}=0.033L$ (similar to Fig.~\ref{fig:cost_function_shape}b) and $\hat{v}_3=L$ and $\eta_3=-1$. The optimization target is set to $S^*_{shape}=-L$. This can be set arbitrarily large, but will increase optimization time and dataset size. The data generation of the $50$ PI² optimization runs take $40$ minutes. An overview of the data generation compared to the other models is shown in Table~\ref{tab:experimental_variations2}.
%as a vector of $\mathbf{e}_2$ components of $p_1$ and $p_2$
\begin{table}[t]
    \centering
    \caption{Model Variations}
    \renewcommand{\arraystretch}{1.1}
    \setlength{\tabcolsep}{5pt}
    \begin{tabular}{|c|c|c|c|c|}
        \hline
        \textbf{Model} & \textbf{1P-2D} & \textbf{3P-2D} & \textbf{2P-2D} & \textbf{3P-3D} \\ \hline
        \multicolumn{5}{|c|}{\textbf{PI² Data Generation}} \\ \hline
        %\textbf{Trajectory dimensions} & 2D & 2D & 2D & 3D \\ \hline
        \textbf{Size of $\boldsymbol{\theta}$} (N) & 10 & 10 & 20 & 60 \\ \hline
        $\sigma^-$ & 0.0003 & 0.0007 & 0.006 & 0.0015 \\ \hline
        $\sigma^+$ & 0.05 & 0.13 & 0.1 & 0.06 \\ \hline %\textbf{Hyper-parameter} 
        \textbf{Number of PI² runs} & 1 & 50 & 5 & 30 \\ \hline
        %$S_{shape}$ & \ding{51} & \ding{51} & \ding{51} & \ding{51} \\
        $S_{shape}$ in Fig.~\ref{fig:cost_function_shape} & a & b & b & b, c \\
        $S_{scope}$ & 1$\times$ & 3$\times$ & 2$\times$ & 5$\times$ \\
        $S_{acc,0}$, $S_{jerk}$ & \ding{51} & \ding{51} & \ding{51} & \ding{55}  \\ \hline
        \textbf{Target $S^*_{shape}$} & -0.47L & -L & -0.33L & -0.33L \\ \hline
        \textbf{Computation time} & 1m27s & 40m15s & 17m27s & 2h48m \\ \hline
        \multicolumn{5}{|c|}{\textbf{Neural Network Training}} \\ \hline
        \textbf{Training data size} & 199 & 5.1k & 1.0k & 7.4k \\ \hline
        \textbf{Input size $n$}& 1 & 3 & 2 & 3 \\
        \textbf{Hidden layer size} & 1028 & 256$\times$512 & 256$\times$512 & 256$\times$512 \\ 
        \textbf{Output size} & 20 & 20 & 40 & 180 \\ \hline 
        \textbf{Computation time} & 1s & 11s & 4s & 39s \\ \hline
    \end{tabular}
    
    \label{tab:experimental_variations2}
\end{table}

%\begin{equation}
%    S = c_1S_{shape}+c_2S_{scope}+c_3S_{acc,0}+c_4S_{jerk},
%\end{equation}
%where $c_1$

\section{Neural Network Training and Testing}
\label{sec:nn}
A single PI² optimization run generates a unique trajectory at each step $j$ defined by $S_{shape,j}$. This trajectory is labeled by the corresponding forcing term parameters $\boldsymbol{\theta}_j$. When reaching the target $S^*_{shape}$, $J=j^*$ trajectories can be added to the dataset $\mathcal{D}$. In this manner, a single optimization run creates the whole dataset for one-dimensional task parameter vector $\mathcal{T}=[s_1]$, where $s_1=S_{shape}$. More dimensional task parameter spaces will get additional inputs based on $s_2$ and $s_3$ that distinguish different PI² optimization runs $j$. In these cases, the dataset is generated by drawing an equal number of samples from each optimization run using the minimum number of completion steps $J_{min,j}$ across all runs. When a run has more steps than $J_{min,j}$, samples are drawn uniformly to maintain balance. 
The neural network (NN) architecture consists of one or two fully connected hidden layers, the training is performed with the \textit{Adam} optimizer and a learning rate of $0.0005$ using the Mean Squared Error (MSE) loss. The training is implemented with PyTorch~\cite{paszke2019pytorch} and the trained model is saved as a scripted model. As the mapping of the optimal PI² trajectories to the NN model induces inaccuracies, the model performance is evaluated by the $s_1$-error using $1000$ randomly generated task parameters. For obstacle avoidance applications, positive errors are less problematic, as they only create less optimal avoidance. Negative errors, on the other side, can cause collisions. Hence, an offset $o$ is added to the task parameters to mitigate negative errors.

\begin{figure}[t]
    \centering 
    \includegraphics[width=0.49\textwidth]{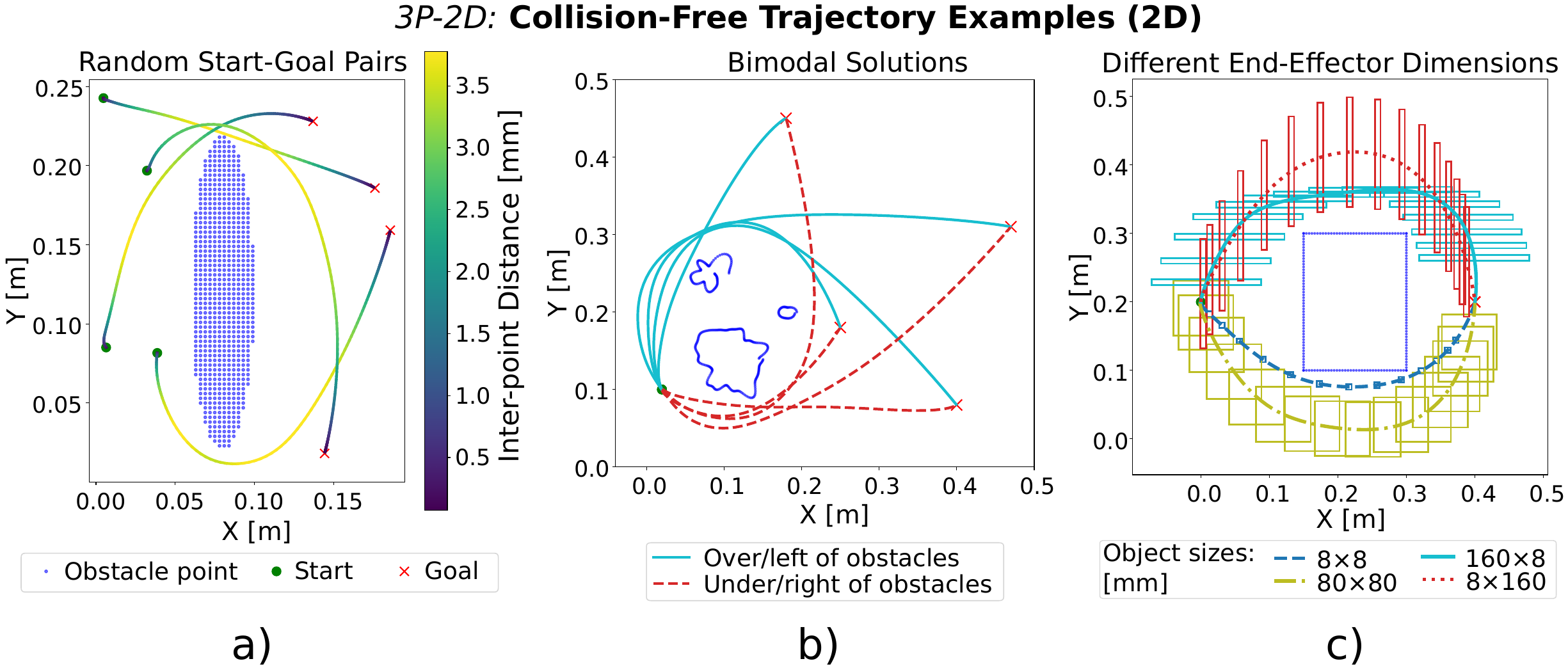}
    \caption{2D avoidance example using the \textit{3P-2D} model with different obstacles (blue): a) near-optimal, smooth trajectories for random start and goal initialization, b) bi-modal solutions, c) considering four different obstacle dimensions.}
    \label{fig:simple_2D_avoidance}
\end{figure}

%\subsection{1P-2D model}
%One PI² optimization run generates $199$ trajectories until reaching the target $S^*_{shape}$, which are mapped in a NN using one hidden layer with $1028$ nodes, chosen empirically. The training is performed for $60$ epochs within $1$ s. The $1000$ random test runs show that an offset $o=0.05L$ results in a $100$\% avoidance success rate.
%\subsection{3P-2D model}
Using the \textit{3P-2D} model example, fifty PI² optimization runs generate $5.1$k trajectories for the NN using two hidden layers with $256$ and $512$ nodes. The training is performed for $100$ epochs within $11$ s. The model is tested with $1000$ random task parameter samples for $\mathcal{T}=[s_1,s_2,s_3]$. An offset \mbox{$o=0.02L$} results in a $100$\% avoidance success rate. It is applied to all three task parameters to increase the avoidance space: $s_1+o$, $s_2-o$, $s_3+o$. A 2D avoidance example illustrates the model's ability to create near-optimal trajectories for four different start-goal pairs that are randomly sampled on either side of an ellipsoidal obstacle (Fig.~\ref{fig:simple_2D_avoidance}a). In Fig.~\ref{fig:simple_2D_avoidance}b, instead of only choosing the shortest path, a second solution is shown, avoiding three obstacles to both sides. An overview of the NN training compared to the other models is shown in Table~\ref{tab:experimental_variations2}.
The flexibility of setting different offsets can also be used for different end-effector or mobile robot dimensions $\mathbf{w}$, e.g. $s_1+\mathbf{w}_{\mathbf{e}_2}$, $s_2-0.5\mathbf{w}_{\mathbf{e}_1}$ and $s_3+0.5\mathbf{w}_{\mathbf{e}_1}$. In Fig.~\ref{fig:simple_2D_avoidance}c, trajectory solutions with different end-effector dimensions are presented.

\section{Experiments}
\label{sec:sim}
The effectiveness of the proposed approach is presented in four 3D experiments with a UR5e manipulator and realsense camera in simulation using ROS 2 Humble and Gazebo 11 Classic. We carry out experiments in a variety of tasks: a pick-and-drop, an insertion, and a 3D avoidance experiment that are inspired by~\cite{tekden2023neural}, as well as a stacking experiment (see Fig.~\ref{fig:fig1}). Finally, the \textit{3P-2D} model is applied in a real robot experiment.

%\begin{table*}[t]
%    \centering
%    \caption{Variations of experimental setups.}
%    \begin{tabular}{|c|c|c|c|c|c|c|}
%        \hline
%        \textbf{Task} & \textbf{PI² cost function sets} & \textbf{# PI² optimizations} & \textbf{Training data size} & \textbf{NN inputs} & \textbf{NN outputs}  & \textbf{Point cloud detection} \\ \hline
%        \textbf{Pick-and-drop} & 0.8 & 0.8 & 0.8 & 0.8 & 0.8 & 0.8 \\ \hline
%        \textbf{Pick-and-drop \& Stacking} & 0.8 & 0.8 & 0.8 & 0.8 & 0.8 & 0.8 \\ \hline
%        \textbf{Insertion}  & 0.8 & 0.8 & 0.8 & 0.8 & 0.8 & 0.8 \\ \hline
%        \textbf{3D avoidance}  & 0.8 & 0.8 & 0.8 & 0.8 & 0.8 & 0.8 \\ \hline   
%    \end{tabular}   
%    \label{tab:pixel_sizes}
%\end{table*}

\subsection{Robot Control}
The initial DMP parameters are obtained from a single demonstration $\textbf{x}_D(t)$. The length $L$ of $\textbf{x}_D(t)$ is used to scale the DMP parameters $\boldsymbol{\theta}$ to maintain the same trajectory shape. The current end-effector position is sent to the detection node that detects the goal position and the task parameters (see Fig.~\ref{fig:fig1}). The neural network model returns the respective $\boldsymbol{\theta}$ that are set in the DMP and allows generating a collision-free Cartesian trajectory. This trajectory is translated into a joint trajectory using inverse kinematics (IK) (via MoveIt~\cite{moveit} or the Kinenik library\footnote{https://gitioc.upc.edu/robots/kinenik}), and sent to the robot using a ROS 2 joint trajectory action. 

\subsection{Point Cloud Detection}
\label{sec:pcd}
To perform the experiments automatically in different scenarios, point cloud data is used to detect the goal object and the obstacles. The goal position of the trajectory can be derived by detecting the goal object, the required trajectory shape is characterized by the task parameters that depend on the model, and can be derived when knowing the initial and goal positions of the end-effector in world coordinates. 
%\subsection{Point cloud pre-processing}
%
To this end, the point cloud data is first preprocessed using Open3D~\cite{Zhou2018open3d}. To reduce computation times, the point cloud is downsampled with a voxel size of $0.01$ m. Then, the point cloud is transformed into world coordinates given the camera pose. After this, the current end-effector position can be determined in the point cloud and the links of the robot manipulator can be removed from the point cloud to avoid false obstacle detections. Finally, the points belonging to the floor are removed as well, by considering only points with a $Z$ coordinate above the floor, which leaves the obstacles and the goal object. 
The obstacle region is assumed to be in the middle within $X = [-0.1,0.1]$ m and $Y>0.25$ m. The goal region is assumed to be on the opposite side to the initial end-effector position. Goal objects are detected using the 2D density-based clustering method DBSCAN~\cite{ester1996density} implemented with scikit-learn~\cite{pedregosa2011scikit-learn}. Points to be considered in a cluster can have a maximum distance between points of \mbox{$0.02$ m} and the cluster requires at least ten points. Finally, the size of the found clusters are matched with the $X-Y$ dimensions of the goal object, allowing $0.02$ m tolerance. The goal position is returned as the center point of the 2D cluster and the maximum $Z$ position of the points that belong to the cluster.

\begin{figure}[t]
    \centering 
    \includegraphics[width=0.49\textwidth]{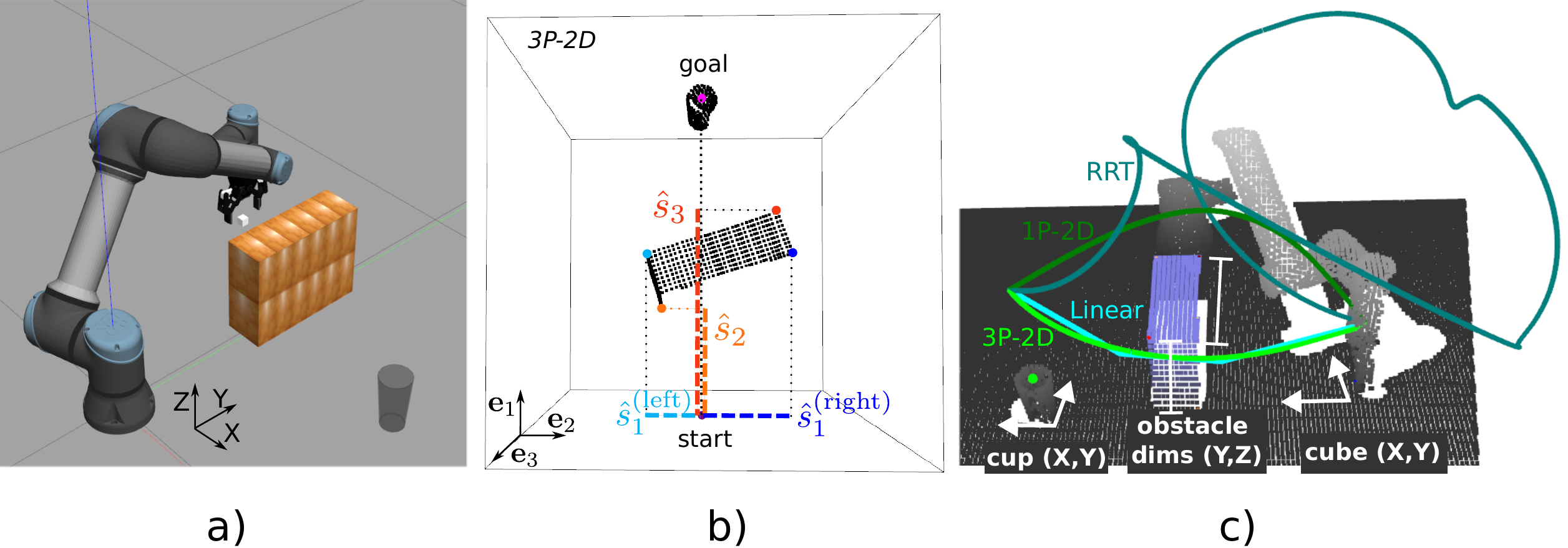}
    \caption{Pick-and-Drop experiment: a-b) task parameter derivation with \mbox{\textit{3P-2D}}, and c) experiment setup and comparison to \textit{1P-2D} and baselines.}
    \label{fig:obstacle_detection_experiment}
\end{figure}

In the example with the \textit{3P-2D} model, three task parameters must be derived, capturing the position and dimensions of the obstacle. Figure~\ref{fig:obstacle_detection_experiment}b shows an example in a rotated coordinate frame with the preliminary task parameters $\hat{\mathcal{T}}=[\hat{s}_1,\hat{s}_2, \hat{s}_3]$ that create a collision-free trajectory for a point mass. The final task parameters $\mathcal{T}$ consider the end-effector dimensions $\mathbf{w}$ as shown previously in Fig~\ref{fig:simple_2D_avoidance}c. However, in this 3D experiment, three solutions for $s_1$ are considered by finding the points with the maximum $\mathbf{e}_2$ (right) and $\mathbf{e}_3$ (up) values as well as the minimum $\mathbf{e}_2$ (left) value, creating multi-modal solutions. 
%Here, the complete obstacle space is evaluated based on the point that has the minimum distance in $\mathbf{e}_1$ to the start and goal, representing the initial task parameters $\hat{s}_2$ and $\hat{s}_3$ respectively. The required trajectory height $s_1$ is derived in three directions,   

\subsection{Pick-and-Drop}
For this experiment the models \textit{1P-2D} and \textit{3P-2D} are compared with each other and with two baselines. The \textit{Linear} baseline combines three linear segments where the two intermediate waypoints are provided by the {3P-2D} detection. Our models and the \textit{Linear} baseline create trajectories in Cartesian space. The second baseline plans trajectories in joint space using RRT-Connect~\cite{kuffner2000rrt} implemented in MoveIt\cite{moveit}. The Cartesian trajectories are also executed via MoveIt to consider the same workspace and joint acceleration constraints.

\begin{figure*}[t]
    \centering 
    \includegraphics[width=0.98\textwidth]{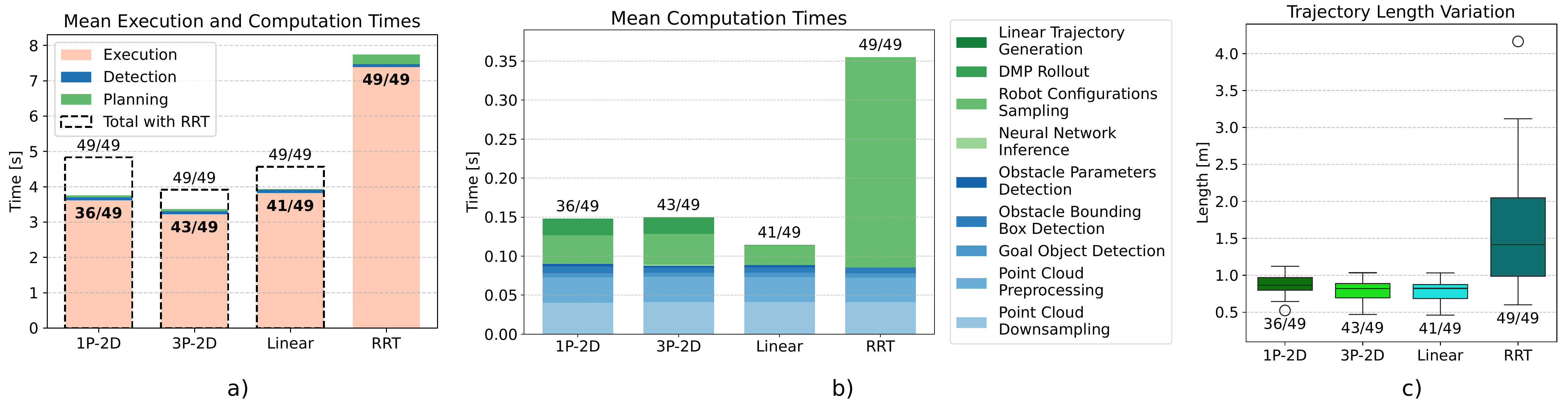}
    \caption{Comparison of our \textit{1P-2D} and \textit{3P-2D} models to two baselines: \textit{Linear} and \textit{RRT}. a) Computation time for detection and planning and execution time. b) Detailed contributions to computation time for detection and planning. c) Variations of the trajectory lengths.}
    \label{fig:exp_comparison}
\end{figure*}

The pick-and-drop experiment requires picking and carrying a small cube to the other side of an obstacle and into a cup (Fig.~~\ref{fig:obstacle_detection_experiment}a). The position of the small cube and the cup are sampled randomly inside two workspaces with dimensions $0.4 \times 0.4$ m on opposite sides of the obstacle (see an example in Fig.~\ref{fig:obstacle_detection_experiment}c). The dimensions of the obstacle are defined by the arrangement of up to $20$ blocks, each of size \mbox{$0.1 \times 0.05 \times 0.12$ m}. Between one and seven blocks are selected randomly along the $Y$ axis ($n_Y \in [1,7]$) and between one and four blocks are stacked along the $Z$ axis ($n_Z \in [1,4]$). The $Y$ location of the blocks depends on $n_Y$, such that the obstacle arrangement is centered at $Y=0.5$ m, while $X=0$ is constant for all blocks.
%
%~\ref{fig:exp_pcd_trajs}
%\begin{figure}[t]
%    \centering 
%    \includegraphics[width=0.49\textwidth]{graphics/pickndrop_pcd_trajs.pdf}
%    \caption{Pick-and-drop setup: The robot manipulator grasps a cube on the right side. The goal is detected on the left and a collision-free trajectories are executed. The configuration is sampled randomly within a given range. Two examples are shown in the figure (a) and (b).}
%    \label{fig:exp_pcd_trajs}
%\end{figure}
%
For the comparison, $50$ random configurations are sampled. 
% One configuration starts in a collision and is therefore not feasible for any method. 
The execution time, planning time, detection time and trajectory length are evaluated for each method. If one of the methods that generate Cartesian trajectories does not find a collision-free trajectory, it falls back to \textit{RRT}. In Fig.~\ref{fig:exp_comparison}, the number of successful runs per method is indicated in bold font. After falling back to \textit{RRT}, all runs are completed successfully. The outlined block with dashed lines shows the sum of execution, detection and planning times for all runs, including the \textit{RRT} fallbacks. 

%Each \textit{RRT} fallback is computed as the sum of the method's detection and planning time at the specific run, and the average planning and execution time of the \textit{RRT} method. This is chosen to make the comparison of the three Cartesian methods independent from the high variance of the \textit{RRT}'s planning and execution times (see Fig.~\ref{fig:exp_variation_comparison}).

In Fig.~\ref{fig:exp_comparison}a, our DMP-based models are the fastest in terms of combined computation times for detection and planning, and the execution times, considering only runs in which they found a feasible solution. The \textit{RRT} requires the longest execution and planning times on average. However, it finds a solution in all $49$ feasible experiment runs. The \textit{1P-2D} method does not find a feasible solution in $13$ runs and requires falling back to \textit{RRT}. This results in an accumulated time that exceeds the \textit{Linear} approach. For comparison, the accumulated mean \textit{RRT} time is added when the Cartesian methods cannot find a feasible solution. In those cases, the planning times for generating a Cartesian trajectory are included. As shown in more detail in Fig.~\ref{fig:exp_comparison}b, this adds less than $0.1$ s to the total runtime. The \textit{Linear} solution has the smallest planning time, but requires more time to execute the trajectory, as the abrupt changes of direction between the linear sections would create large accelerations that the MoveIt controller dampens.  
    %In (b), the detailed detection and planning operations are listed including their respective computation times, showing that 
The robot configuration sampling of the \textit{RRT} takes the most time, as well as sampling the Cartesian trajectory collision checks. On the other hand, the neural network inference and the linear trajectory generation contribute marginally. The detection times are similar for all methods. The \textit{RRT} does not require computing the obstacle parameters, which is a marginal effort. The initial downsampling of the point cloud takes the most time in the detection part.  
Figure~\ref{fig:exp_comparison}c illustrates large variations of the trajectory lengths generated by the \textit{RRT}. The large variations increase the average for the \textit{RRT} in the comparison, however, even the best performing runs cannot outperform the Cartesian methods. Here, the \textit{1P-2D} model creates longer trajectories than the \textit{Linear} method, but they can be executed faster due their smooth nature. 

The \textit{3P-2D} model combines short trajectory lengths and smooth execution, resulting in the best overall performance. Its combined planning and execution times are $16$\% faster on average than the \textit{Linear} solutions. In general, the Cartesian methods clearly outperform \textit{RRT}, however, \textit{RRT} is the only method that finds a solution to all feasible runs. The combination of the \textit{3P-2D} with \textit{RRT} as fallback option is therefore the best choice for a fast and reliable operation.

%\begin{figure}[t]
%    \centering 
%    \includegraphics[width=0.49\textwidth]{graphics/pickndrop_variation_comp.pdf}
%    \caption{Comparison of the variations of the trajectory length, execution time and planning time.}
%    \label{fig:exp_variation_comparison}
%\end{figure}

%consider explaining the use of different controller (without moveit)
\subsection{Stacking, Insertion and 3D Avoidance}
To show the scalability and transferability of our approach, three additional experiments are conducted.
The stacking experiment involves a sequence of two pick-and-place operations to stack two cubes on a third, avoiding similar obstacles as in the pick-and-drop experiment. To this end, three trajectories are generated using the same \textit{3P-2D} model from the pick-and-drop experiment. The initial point cloud, the three generated trajectories and a snapshot from the simulation right before placing the second cube are shown in Fig~\ref{fig:fig1}. Unlike the pick-and-drop experiment, the goal object must be approached closely, to achieve a precise tower, and the Z coordinate of the goal location changes in the stacking experiment. Five stacks with different initial cube locations are performed successfully.

The insertion experiment comprises two objects, one peg and a box with a hole that allows $0.005$ m insertion tolerance. Two trajectories are generated, one to the pick location of the peg and another one to the insertion location of the hole. Particularly, the insertion requires a precise vertical descend at the end of the trajectory. The \textit{2P-2D} model is trained for that purpose using $S_{scope}$ with an upper bound very close to the goal, as well as a $S_{shape}$ where $p_2$ is constant and very close to the goal. Five PI² optimizations with different $p_1$ locations are performed to create the dataset. For each of $20$ experiment runs, the positions of the peg and box are sampled randomly inside the workspace. All runs succeed with the insertion. In two cases, an initial collision between peg and box can be observed at the beginning of the insertion process. % (compare to Fig.~\ref{fig:cost_function_shape}b)

For the 3D avoidance experiment, the \textit{3P-3D} model is trained to generate a three-dimensional trajectory that avoids obstacles in a pre-defined way. Two outer walls of varying heights require avoidance above, while a gap in between with varying $Y$ position and $0.02$ m tolerance requires the end-effector to pass through. Unlike the other experiments, this one has constant start and goal positions. The three tasks parameters describe the heights of the walls ($s_1,s_2)$ and the position of the gap ($s_3$). In $20$ experiment runs with random gap location and the same wall configurations as in the training, all but one are successful. The one failed experiment collided inside the gap. This shows the challenge with gaps as no offset can be set to improve the performance. The related cost factor $C_{scope}=1$ (eq.~\ref{eq:cost_scope}) could be increased as the current value does not completely keep the trajectory inside the gap (see Fig.~\ref{fig:cost_function_shape}c).
More details about these models and experiments are available online\footnote{\url{https://github.com/DominikUrbaniak/obst-avoid-dmp-pi2}}.

\subsection{Real Robot Experiments}
\label{sec:real}
We show here the transferability of the \textit{3P-2D} model to experiments with a real robot setting (Fig.~\ref{fig:madar_pickndrop}).
%manipulator and a stereo camera (Fig.~\ref{fig:madar_pickndrop}). 
% \subsubsection{Utilized hardware}
\subsubsection{Hardware setup}
\begin{figure}[t]
    \centering 
    \includegraphics[width=0.49\textwidth]{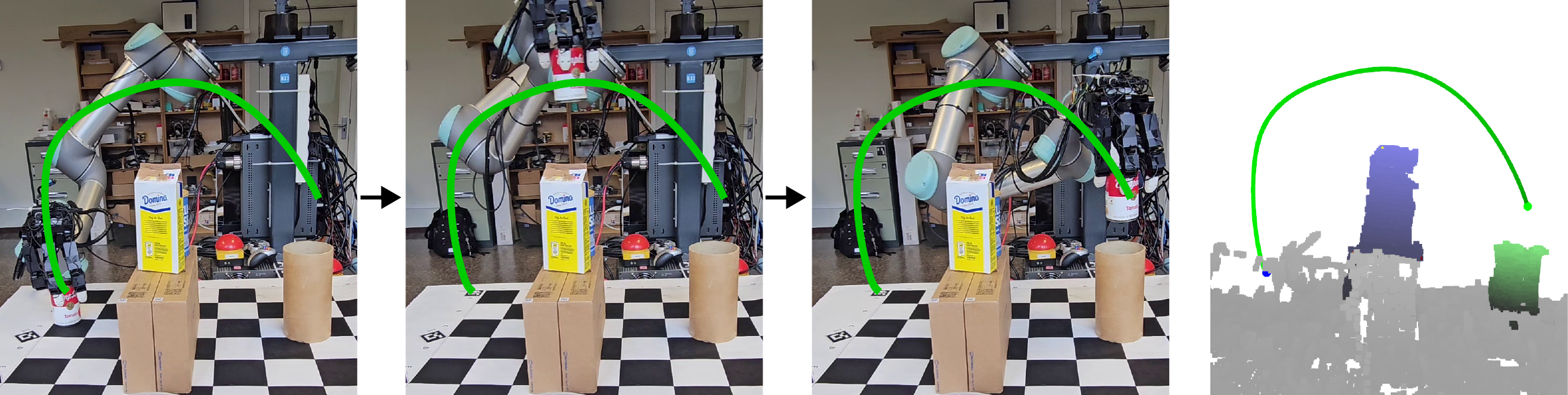}
    \caption{Real robot manipulator and stereo camera. Three snapshots of a real pick-and-drop experiment and the corresponding point cloud image with the generated DMP trajectory (green line). In the point cloud, blue points mark the obstacle region and green points mark the goal region.}
    \label{fig:madar_pickndrop}
\end{figure}

%\paragraph{}{Mobile manipulator - MADAR}
\textbf{Mobile manipulator MADAR:}
In the experiments, one UR5 arm of the dexterous dual-arm omnidirectional mobile manipulator (MADAR) is used~\cite{suarez2018development}. Avoidance experiments are conducted without a hand and for the pick-and-drop experiment an Allegro hand is used. The camera on the MADAR is used to calibrate the pose of the external camera rc\_visard (described below) by detecting the same markers on a table. 
%
%\paragraph{Stereo camera - rc\_visard}
\textbf{Stereo camera rc\_visard:}
As external camera, the industrial stereo camera by Roboception GmbH (rc\_visard) is utilized. It generates a stable and smooth point cloud without modifying the workspace and keeping the same chessboard surface with the markers on the table (see Fig.~\ref{fig:madar_pickndrop}). 
%\paragraph{Private 5G network}
\textbf{Private 5G network:}
A private 5G network is used to communicate the generated collision-free trajectory to the robot controller. It has an open-source 5G core (Open5GS), a Node-H Askey 5G SCE2120 small cell in the N77 band and four Teltonika RUTX50 routers as receivers. 
\subsubsection{Experimental setup and results}
Starting with the robot end-effector on one side of an obstacle, the goal object is placed on the other side of the obstacle, such that it is reachable and a collision-free trajectory exists. Subsequently, the point cloud areas of the goal object and the obstacle are processed on the edge computer, receiving the point cloud directly from the camera. The collision-free Cartesian trajectory is computed and interpolated based on a desired execution duration, and an IK service on the MADAR is called via 5G to convert it into a joint trajectory. Finally, the joint trajectory is sent to the MADAR controller and executed.  
There are two main differences compared to the simulated experiment. First, the UR5 is mounted on the MADAR instead of on the ground. The model generates the same Cartesian trajectory, but the joint trajectories are different. This shows that the approach is model-free and can be used with any robot manipulator. Secondly, besides boxes and a cup, various household objects are used as obstacles and three goal objects of different heights are utilized. This shows that any object can be avoided or serve as goal object as long as the camera can create a point cloud of it. 
Three main observations can be drawn from the real experiment: i) the trajectory execution is smooth, even when reducing the execution time to 3 s; ii) the extent to which the grasped object alters the collision space is difficult to detect, however, important when creating near-optimal trajectories. Our experiment cannot consider that automatically, but manually adjusting the trajectory shape with the offset $o$ handles that challenge; iii) inaccuracies are mainly caused by the goal detection, which can cause pick-and-drop failures when tolerances are tight.

\section{Discussion and Future Work}
\label{sec:discussion}
\begin{figure}[t]
    \centering 
    \includegraphics[width=0.49\textwidth]{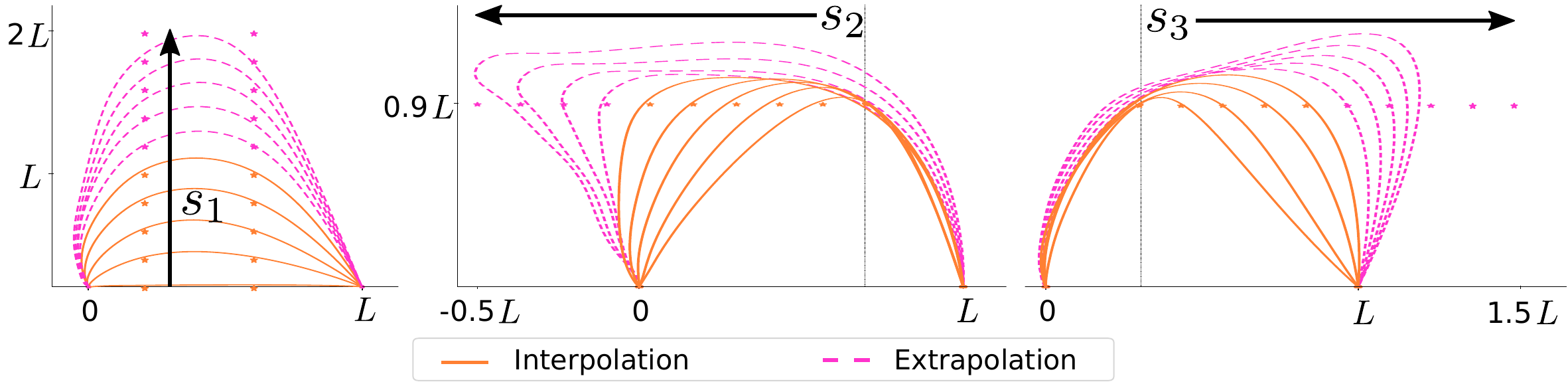}
    \caption{Extrapolation ability of \textit{3P-2D} for each of the three task parameter.}
    \label{fig:extrapolation}
\end{figure}

Three limitations of the approach are presented below and how they could be addressed in future work.
\textbf{Error scaling:} The scaling feature of the DMPs increases the generalization abilities of the approach. However, it also scales the error, complicating the use with tight tolerances, like insertion or navigating through a gap. As detection inaccuracies further complicate these types of applications, we plan to approach this limitation in the future by using the DMP's dynamic adaptation ability.
\textbf{Engineered point cloud processing:} Currently, the task parameters are derived automatically from a point cloud, after engineered processing. Learning the task parameter selection directly from a point cloud could be promising and might even improve the optimality of the \textit{3P-2D} solutions when there are multiple obstacles present, which are currently considered by a bounding box over all objects. 
%
%The presented approach relies on trajectory shapes that continuously evolve along the task parameters. More complex trajectory shapes require more training effort and are more difficult to be reproduced accurately with the NN model. In future work, these challenges could be addressed by leveraging the DMP's dynamic adaptability. With a suitable controller, the DMP can be executed step-by-step, allowing the adjustment of the goal position and forcing terms parameters on-the-fly. These could be predicted by an additional policy-free RL agent using the point cloud as input.
%to predict the right task parameters.The forcing term parameters can also be changed dynamically. However, the effect on the trajectory will depend on the current state of the execution of the initial trajectory. Hence, the current task parameters lose their intuitive description of the detected obstacle. Instead, an additional policy-free RL agent could be trained using the point cloud as input to predict the right task parameters. % that generate the desired dynamic trajectory perturbation.
%
\textbf{Extrapolation abilities:} Even though the training data is collected in a user-defined range, the network is able to handle task parameters outside that range. Fig.~\ref{fig:extrapolation} shows that the generated trajectories keep the desired evolution for each task parameter, however, the accuracy decreases. In future work, the parameter mapping could be improved, possibly by replacing the deterministic neural network with a generative model, which shows better extrapolation performance in a 1D example~\cite{pervez2018learning}.  

\section{Conclusion}
We presented a method that generates collision-free trajectories online within $0.2$ s, after two minutes to less than three hours of offline training, based on a single artificially-generated demonstration. The application to four different scenarios was presented, showing generalization to varying goal locations and obstacle types and sizes. Its smooth and near-optimal trajectories are executed on average at least $16$\% faster than the \textit{Linear} and RRT-Connect baselines, given the same joint acceleration limits. The required task parameters are automatically derived from a point cloud, which was shown in simulated and real experiments.
%\vfill

%\balance
\renewcommand\refname{Bibliography}
  \addcontentsline{toc}{section}{Bibliography}
  \bibliographystyle{IEEEtran}
  
  \bibliography{references}

\end{document}